\documentclass[11pt,a4paper]{article}
\usepackage[hyperref]{naaclhlt2019}
\usepackage{times}
\usepackage{latexsym}

\usepackage{url}

\usepackage{comment}
\usepackage{array, booktabs}
\usepackage{amsmath,amssymb,bm}
\usepackage{graphicx}
\usepackage{multirow}

\aclfinalcopy 

\newcommand{\argmax}{\mathop{\rm argmax}\limits}
\newcommand{\argmin}{\mathop{\rm argmin}\limits}

\title{Multimodal Machine Translation with Embedding Prediction}

\author{
    Tosho Hirasawa \and
    Hayahide Yamagishi \and
    Yukio Matsumura \and
    Mamoru Komachi \\
    Tokyo Metropolitan University \\
    \texttt{\{hirasawa-tosho@ed., yamagishi-hayahide@ed.,} \\
    \texttt{ matsumura-yukio@ed., komachi@\}tmu.ac.jp}
}

\date{}

\begin{document}
\maketitle
\begin{abstract}
    Multimodal machine translation is an attractive application of neural machine translation (NMT).
    It helps computers to deeply understand visual objects and their relations with natural languages.
    However, multimodal NMT systems suffer from a shortage of available training data, 
    resulting in poor performance for translating rare words.
    In NMT, pretrained word embeddings have been shown to improve NMT of low-resource domains, 
    and a search-based approach is proposed to address the rare word problem.
    In this study, we effectively combine these two approaches in the context of multimodal NMT
    and explore how we can take full advantage of pretrained word embeddings to better translate rare words.
    We report overall performance improvements of 1.24 METEOR and 2.49 BLEU
    and achieve an improvement of 7.67 F-score for rare word translation.
\end{abstract}

\section{Introduction}

In \textbf{multimodal machine translation}, a target sentence is translated from a source sentence 
together with related nonlinguistic information such as visual information.
Recently, neural machine translation (NMT) has superseded traditional statistical machine translation owing to the introduction of the attentional encoder-decoder model, in which machine translation is treated as a sequence-to-sequence learning problem and is trained to pay attention to the source sentence while decoding \cite{bahdanau2015jointly}.

Most previous studies on multimodal machine translation are classified into two categories: visual feature adaptation and data augmentation.
In visual feature adaptation, multitask learning \cite{elliott2017imagination} and feature integration architecture \cite{caglayan2017liumcvc,calixto2017doubly} are proposed to improve neural network models.
Data augmentation aims to deal with the fact that the size of available datasets for multimodal translation is quite small.
To alleviate this problem, parallel corpora without a visual source \cite{elliott2017imagination,gronroos2018memad} and pseudo-parallel corpora obtained using back-translation \cite{helcl2018cuni} are used as additional learning resources.

Due to the availability of parallel corpora for NMT, \newcite{qi2018when} suggested that initializing the encoder with pretrained word embedding improves the translation performance in low-resource language pairs.
Recently, \newcite{kumar2018von} proposed an NMT model that predicts the embedding of output words and searches for the output word instead of calculating the probability using the softmax function.
This model performed as well as conventional NMT, 
and it significantly improved the translation accuracy for rare words.

In this study, we introduce an NMT model with embedding prediction for multimodal machine translation
that fully uses pretrained embeddings to improve the translation accuracy for rare words. 

The main contributions of this study are as follows:
\begin{enumerate}
  \item We propose a novel multimodal machine translation model with embedding prediction 
        and explore various settings to take full advantage of word embeddings.
  \item We show that pretrained word embeddings improve the model performance, especially when translating rare words.
\end{enumerate}

\section{Multimodal Machine Translation with Embedding Prediction}

We integrate an embedding prediction framework \cite{kumar2018von} with the multimodal machine translation model and take advantage of pretrained word embeddings.
To highlight the effect of pretrained word embeddings and embedding prediction architecture, 
we adopt IMAGINATION \cite{elliott2017imagination} as a simple multimodal baseline.

IMAGINATION jointly learns machine translation and visual latent space models.
It is based on a conventional NMT model for a machine translation task.
In latent space learning, a source sentence and the paired image are mapped closely in the latent space.
We use the latent space learning model as it is, except for the preprocessing of images.
The models for each task share the same textual encoder in a multitask scenario.

The loss function for multitask learning is the linear interpolation of loss functions for each task.
\begin{equation}
  J = \lambda J_\mathrm{T}(\theta,\phi_\mathrm{T}) + (1 - \lambda)J_\mathrm{V}(\theta,\phi_\mathrm{V})    
\end{equation}
where $\theta$ is the parameter of the shared encoder; $\phi_\mathrm{T}$ and $\phi_\mathrm{V}$ are parameters of the machine translation model and latent space model, respectively; and
$\lambda$ is the interpolation coefficient\footnote{We use $\lambda=0.01$ in the experiment.}.

\subsection{Neural Machine Translation with Embedding Prediction}

The machine translation part in our proposed model is an extension of \newcite{bahdanau2015jointly}.
However, instead of using the probability of each word in the decoder, it searches for output words based on their similarity with word embeddings.
Once the model predicts a word embedding, its nearest neighbor in the pretrained word embeddings is selected as the system output.
\begin{eqnarray}
  \bm{\hat{e}_j} &=& \tanh(\bm{W_\mathrm{o}} \bm{s_j} + \bm{b_\mathrm{o}}) \\
  \hat{y}_j &=& \argmin_{w \in \mathcal{V} }\{d(\bm{\hat{e}_j}, \bm{\mathrm{e}}(w))\}
\end{eqnarray}
where $\bm{s_j}$, $\bm{\hat{e}_j}$, and $\hat{y}_j$ are the hidden state of the decoder, predicted embedding, 
and system output, respectively, for each timestep $j$ in the decoding process.
$\bm{\mathrm{e}}(w)$ is the pretrained word embedding for a target word $w$.
$d$ is a distance function that is used to calculate the word similarity.
$\bm{W_\mathrm{o}}$ and $\bm{b_\mathrm{o}}$ are parameters of the output layer.

We adopt margin-based ranking loss \cite{lazaridou2015hubness} as the loss function of the machine translation model.
\begin{equation}
\begin{split}
  J_\mathrm{T}(\theta, \phi_\mathrm{T}) = \sum_j^M \max \{0, \gamma + d(\bm{\hat{e}_j}, \bm{\mathrm{e}}(w_{j}^{-})) \\
  \quad - d(\bm{\hat{e}_j}, \bm{\mathrm{e}}(y_j)) \}
\end{split}
\end{equation}
\begin{equation}
    w_{j}^{-} = \argmax_{w \in \mathcal{V}} \{ d(\bm{\hat{e}_j}, \bm{\mathrm{e}}(w)) 
    - d(\bm{\hat{e}_j}, \bm{\mathrm{e}}(y_j))
\end{equation}
where $M$ is the length of a target sentence and $\gamma$ is the margin\footnote{We use $\gamma=0.5$ in the experiment.}.
$w_{j}^{-}$ is a negative sample that is close to the predicted embedding and far from the gold embedding as measuring using $d$.

Pretrained word embeddings are also used to initialize the embedding layers of the encoder and decoder, and the output layer of the decoder.
The embedding layer of the encoder is updated during training, and the embedding layer of the decoder is fixed to the initial value.

\subsection{Visual Latent Space Learning}

The decoder of this model calculates the average vector over the hidden states $\bm{h_i}$ in the encoder and maps it to the final vector $\bm{\hat{v}}$ in the latent space.
\begin{equation}
    \bm{\hat{v}} = \tanh(\bm{W_\mathrm{v}} \cdot \frac{1}{N} \sum_i^N \bm{h_i} )
\end{equation}
where $N$ is the length of an input sentence and $\bm{W_\mathrm{v}} \in \mathbb{R}^{N*M}$ is learned parameter of the model.

We use max margin loss as the loss function; it learns to make corresponding latent vectors of a source sentence and the paired image closer.
\begin{equation}
  J_\mathrm{V}(\theta,\phi_\mathrm{V}) = \sum_{\bm{v'} \neq \bm{v}} \max \{ 0, \alpha + d(\bm{\hat{v}}, \bm{v'}) 
  - d(\bm{\hat{v}}, \bm{v}) \}
\end{equation}
where $\bm{v}$ is the latent vector of the paired image; $\bm{v'}$, the image vector for other examples; and $\alpha$, the margin that adjusts the sparseness of each vector in the latent space\footnote{We use $\alpha=0.1$ in our experiment.}.

\section{Experiment}

\subsection{Dataset}

We train, validate, and test our model with the Multi30k \cite{elliott2016multi30k} dataset published in the WMT17 Shared Task.

We choose French as the source language and English as the target one.
The vocabulary size of both the source and the target languages is 10,000.
Following \newcite{kumar2018von}, byte pair encoding \cite{sennrich2016bpe} is not applied.
The source and target sentences are preprocessed with lower-casing, tokenizing and normalizing the punctuation.

Visual features are extracted using pretrained ResNet \cite{he2016deep}.
Specifically, we encode all images in Multi30k with ResNet-50 and pick out the hidden state in the pool5 layer as a 2,048-dimension visual feature.
We calculate the centroid of visual features in the training dataset as the bias vector
and subtract the bias vector from all visual features in the training, validation and test datasets.

\subsection{Model}

The model is implemented using nmtpytorch toolkit v3.0.0\footnote{https://github.com/toshohirasawa/nmtpytorch-emb-pred} \cite{caglayan2017nmtpy}.

The shared encoder has 256 hidden dimensions, and therefore the bidirectional GRU has 512 dimensions.
The decoder in NMT model has 256 hidden dimension.
The input word embedding size and output vector size is 300 each.
The latent space vector size is 2,048.

We used the Adam optimizer with learning rate of 0.0004.
The gradient norm is clipped to 1.0. The dropout rate is 0.3.

BLEU \cite{papineni2002bleu} and METEOR \cite{denkowski2014meteor} are used as performance metrics.
We also evaluated the models using the F-score of each word; 
this shows how accurately each word is translated into target sentences, as was proposed in \newcite{kumar2018von}.
The F-score is calculated as the harmonic mean of the precision (fraction of produced sentences with a word that is in the references sentences) and the recall (fraction of reference sentences with a word that is in model outputs).
We ran the experiment three times with different random seeds and obtained the mean and variance for each model.

To clarify the effect of pretrained embeddings on machine translation,
we also initialized the encoder and decoder of our models with random values instead of pretrained embeddings,
and investigated the effect of fixing decoder embeddings.

\subsection{Word Embedding}

We use publicly available pretrained FastText \cite{bojanowski2016enriching} embeddings \cite{grave2018learning}.
These word embeddings are trained on Wikipedia and Common Crawl using the CBOW algorithm, and the dimension is 300.

The embedding for unknown words is calculated as the average embedding over words 
that are a part of pretrained embeddings but are not included in the vocabularies.
Both the target and the source embeddings are preprocessed according to \newcite{mu2018allbutthetop}, 
in which all embeddings are debiased to make the average embedding into a zero vector 
and the top five principal components are subtracted for each embedding.

\section{Results}

Table \ref{test2016} shows the overall performance of the proposed and baseline models.
Compared with randomly initialized models, 
our model outperforms the text-only baseline by +2.49 BLEU and +1.24 METEOR, 
and the multimodal baseline by +2.31 BLEU and +1.09 METEOR, respectively.
While pretrained embeddings improve NMT/IMAGINATION models as well,
the improved models are still beyond our model.

\begin{table}[tp]
    \begin{center}
        \begin{tabular}{l|ccc} \toprule
                        & val   & \multicolumn{2}{c}{test}  \\
            Model       & BLEU  & BLEU          & METEOR        \\ \midrule[0.08em]
            NMT         & 50.83 & 51.00$\pm$.37 & 42.65$\pm$.12 \\
             \small $+$ pretrained & 52.05 & 52.33$\pm$.66 & 43.42$\pm$.13 \\
            IMAG+       & 51.03 & 51.18$\pm$.16 & 42.80$\pm$.19 \\
             \small $+$ pretrained  & 52.40 & 52.75$\pm$.25 & 43.56$\pm$.04 \\ \midrule
            Ours        & \textbf{53.14} & \textbf{53.49$\pm$.20} & \textbf{43.89$\pm$.14} \\ \bottomrule
        \end{tabular}{}
    \end{center}
    \caption{
        Results on Multi30k validation and test dataset.
        NMT denotes the text-only conventional NMT model \cite{bahdanau2015jointly}
        and IMAG+ denotes our reimplementation of the IMAGINATION \cite{elliott2017imagination} model. 
        ``+ pretrained'' models are initialized with pretrained embeddings.
    }
    \label{test2016}
\end{table}

Table \ref{model_comparison} shows the results of ablation experiments of the initialization and fine-tuning methods.
The pretrained embedding models outperform other models by up to +2.77 BLEU and +1.37 METEOR.

\begin{table}[tp]
    \begin{center}
        \begin{tabular}{lll|cc} \toprule
            Encoder  & Decoder  & Fixed & BLEU  & METEOR \\ \midrule[0.08em]
            fasttext & fasttext & Yes   & \textbf{53.49} & \textbf{43.89}  \\ \midrule
            random   & fasttext & Yes   & 53.22          & 43.83  \\
            fasttext & random   & No    & 51.53          & 43.07  \\
            random   & random   & No    & 51.42          & 42.77  \\ \midrule
            fasttext & fasttext & No    & 51.42          & 42.88  \\ 
            random   & fasttext & No    & 50.72          & 42.52  \\ \bottomrule
        \end{tabular}{}
    \end{center}
    \caption{
        Results on test dataset with variations of model initialization and fine-tuning in decoder.
    }
    \label{model_comparison}
\end{table}

\section{Discussion}

\paragraph{Rare Words}

\begin{figure}[tp]
    \begin{center}
        \includegraphics[width=8cm]{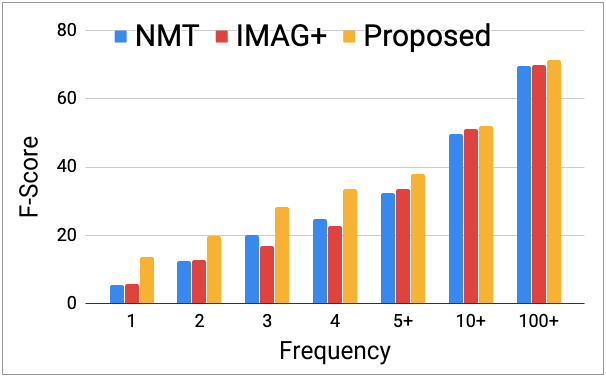}
    \end{center}
    \caption{
        F-score of word prediction per frequency breakdown in training corpus.
    }
    \label{f-score}
\end{figure}

Our model shows a great improvement for low-frequency words.
Figure \ref{f-score} shows a variety of F-score according to the word frequency in the training corpus.
Whereas IMAGINATION improves the translation accuracy uniformly,
our model shows substantial improvement for rare words.

\paragraph{Word Embeddings}

Furthermore, we found that decoder embeddings must be fixed to improve multimodal machine translation with embedding prediction.
When we allow fine-tuning on the embedding layer, the performance drops below the baseline.
It seems that fine-tuning embeddings in NMT with embedding prediction makes the model search for common words more than expected,
thus preventing it from predicting rare words.

More interestingly, using pretrained FastText embeddings on the decoder rather than the encoder improves performance.
This finding is different from \newcite{qi2018when}, in which only the encoder benefits from pretrained embeddings.
Compared with the model initialized with a random value, initializing the decoder with the embedding results in an increase of +1.80 BLEU; in contrast, initializing the encoder results in an increase of only +0.11 BLEU.
This is caused by the multitask learning model that trains the encoder with images and takes it away from what the embedding prediction model wants to learn from the sentences.

\paragraph{Visual Feature}

\begin{table}[tp]
    \begin{center}
        \begin{tabular}{l|ccc} 
            \toprule
                        & val               & \multicolumn{2}{c}{test}              \\
            Model       & BLEU              & BLEU              & METEOR            \\ 
            \midrule[0.08em]
            Ours        & \textbf{53.14}    & \textbf{53.49}    & 43.89             \\
            $-$ Debias    & 52.65             & 53.27             & \textbf{43.91}    \\
            $-$ Images    & 52.97             & 53.25             & \textbf{43.91}    \\ 
            \bottomrule
        \end{tabular}{}
    \end{center}
    \caption{
        Ablation experiments of visual features.
        ``$-$ Debias'' denotes the result without subtracting the bias vector.
        ``$-$ Images'' shows the result of text-only NMT with embedding prediction.
    }
    \label{image_ablation}
\end{table}

We also investigated the effect of images and its preprocessing in NMT with embedding prediction (Table \ref{image_ablation}).
The interesting result is that multitask learning with raw images would not help the predictive model.
Debiasing images is an essential preprocessing for NMT with embedding prediction to use images effectively in multitask learning scenario.

\paragraph{Translation Examples}

\begin{table*}[tp]
    \begin{center}
        \begin{tabular}{l|p{6.2cm}|p{7cm}} \toprule
            \small Image &
            \begin{minipage}{6.2cm}
                \centering
                \includegraphics[height=4.7cm]{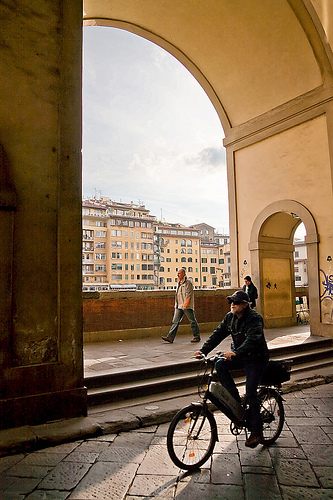}
            \end{minipage} &
            \begin{minipage}{7cm}
                \centering
                \includegraphics[width=7cm]{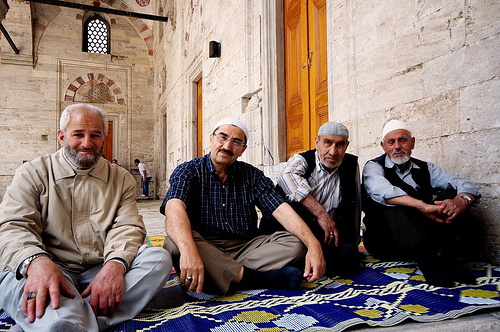}
            \end{minipage} \\ \midrule[0.08em]
            \small Source &
            \small un homme en vélo pédale devant une \textbf{voûte} . & 
            \small quatre hommes , dont trois portent des kippas , sont assis sur un \textbf{tapis} à \textbf{motifs} bleu et vert olive . \\
            \small Reference &
            \small a man on a bicycle pedals through an \textbf{archway} . & 
            \small four men , three of whom are wearing prayer caps , are sitting on a blue and olive green \textbf{patterned} \textbf{mat} . \\
            \small NMT &
            \small a man on a bicycle pedal past an \textbf{arch} . &
            \small four men , three of whom are wearing aprons , are sitting on a blue and green \textbf{speedo} \textbf{carpet} . \\
            \small IMAG+ &
            \small a man on a bicycle pedals outside a \textbf{monument} . &
            \small four men , three of them are wearing alaska , are sitting on a blue \textbf{patterned} \textbf{carpet} and green green seating . \\
            \small Ours &
            \small a man on a bicycle pedals in front of a \textbf{archway} . &
            \small four men , three are wearing these are wearing these are sitting on a blue and green \textbf{patterned} \textbf{mat} . \\ \bottomrule
        \end{tabular}{}
    \end{center}
    \caption{
        French to English translation examples in the Multi30k test set.
    }
    \label{model_examples}
\end{table*}

In Table \ref{model_examples}, we show French-English translations generated by different models.
In the left example, our proposed model correctly translates ``voûte'' into ``archway'' (occurs five times in the training set), 
Although the baseline model translates it to its synonym having higher frequency (nine times for ``arch'' and 12 times for ``monument'').
At the same time, our outputs tend to be less fluent for long sentences.
The right example shows that our model translates some words (``patterned'' and ``carpet'') more concisely; however, it generates a less fluent sentence than the baseline.

\section{Related Works}
Most studies on multimodal machine translation are divided into two categories: visual feature adaptation and data augmentation.

First, in visual feature adaptation, visual features are extracted using image processing techniques and then integrated into a machine translation model.
In contrast, most multitask learning models use latent space learning as their auxiliary task.
\newcite{elliott2017imagination} proposed the IMAGINATION model that learns to construct the corresponding visual feature from the textual hidden states of a source sentence.
The visual model shares its encoder with the machine translation model; this helps in improving the textual encoder.

Second, in data augmentation, parallel corpora without images are widely used as additional training data.
\newcite{gronroos2018memad} trained their multimodal model with parallel corpora and achieved state-of-the-art performance in the WMT 2018.
However, the use of monolingual corpora has seldom been studied in multimodal machine translation.
Our study proposes using word embeddings that are pretrained on monolingual corpora.

\section{Conclusion}

We have proposed a multimodal machine translation model with embedding prediction
and showed that pretrained word embeddings improve the performance in multimodal translation tasks, especially when translating rare words.

In the future, we will tailor the training corpora for embedding learning, 
especially for handling the embedding for unknown words in the context of multimodal machine translation.
We will also incorporate visual features into contextualized word embeddings.

\bibliography{naaclhlt2019}
\bibliographystyle{acl_natbib}

\end{document}